\documentclass[10pt,journal,compsoc]{IEEEtran}
\usepackage{blindtext}
\usepackage[utf8]{inputenc}
\usepackage[T1]{fontenc}
\usepackage{url}
\usepackage{comment}
\usepackage{booktabs}
\usepackage{amsfonts}
\usepackage{multicol}
\usepackage{nicefrac} 
\usepackage{framed}
\usepackage{microtype}
\usepackage{subcaption}
\usepackage{graphicx}
\usepackage{tikz}
\usepackage{float}
\usepackage{sidecap}
\usepackage{amsmath}
\usepackage{multirow}
\usepackage{balance}
\usepackage[font=small,skip=0pt]{caption}

\let\OLDthebibliography\thebibliography
\renewcommand\thebibliography[1]{
  \OLDthebibliography{#1}
  \setlength{\parskip}{0pt}
  \setlength{\itemsep}{0pt plus 0.1ex}
}
\usetikzlibrary{positioning}

\usepackage[font=small]{caption}
\everymath{\displaystyle}
\title{Multi-Moments in Time: \\Learning and Interpreting Models for\\ Multi-Action Video Understanding}
\author{\large  Mathew Monfort$^{13}$, Bowen Pan$^{1}$, Kandan Ramakrishnan$^{3}$, Alex Andonian$^{1}$, Barry A McNamara $^{1}$,\\ Alex Lascelles $^{1}$, Quanfu Fan$^{23}$, Dan Gutfreund$^{23}$, Rogerio Feris$^{23}$,  Aude Oliva$^{13}$\\
$^{1}$ MIT CSAIL, $^{2}$ IBM Research,  $^{3}$ MIT-IBM Watson AI Lab
}

\begin{document}

\maketitle

\begin{abstract} 
Videos capture events that typically contain multiple sequential, and simultaneous, actions even in the span of only a few seconds.  However, most large-scale datasets built to train models for action recognition in video only provide a single label per video.  Consequently, models can be incorrectly penalized for classifying actions that exist in the videos but are not explicitly labeled and do not learn the full spectrum of information present in each video in training.  Towards this goal, we present the Multi-Moments in Time dataset (M-MiT) which includes over two million action labels for over one million three second videos.   This multi-label dataset introduces novel challenges on how to train and analyze models for multi-action detection. Here, we present baseline results for multi-action recognition using loss functions adapted for long tail multi-label learning, provide improved methods for visualizing and interpreting models trained for multi-label action detection and show the strength of transferring models trained on M-MiT to smaller datasets.
\end{abstract}

\section{Introduction}

\begin{figure}
  \centering
\begin{subfigure}{0.8\linewidth}
  \centering
  \includegraphics[width=\linewidth]{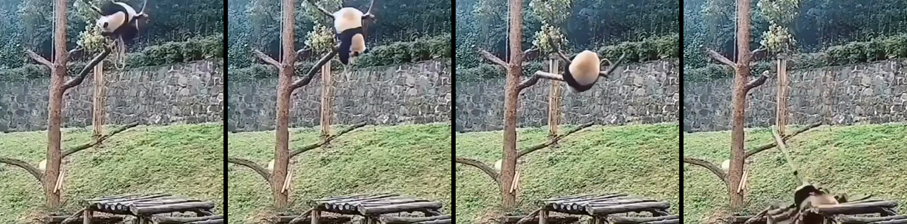}
  \caption{\textbf{Multi-Moments in Time} 2 million action labels for 1 million 3 second videos}
\end{subfigure}
\begin{subfigure}{0.8\linewidth}
  \centering
  \includegraphics[width=\linewidth]{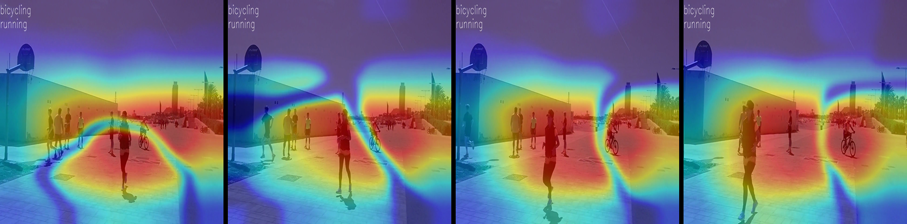}
  \caption{\textbf{Multi-Regions} Localizing multiple visual regions involved in recognizing simultaneous actions, like \emph{running} and \emph{bicycling}}
\end{subfigure}
\begin{subfigure}{0.8\linewidth}
  \centering
  \includegraphics[width=\linewidth]{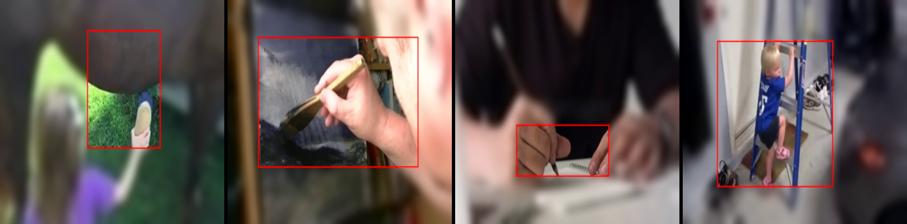}
  \caption{\textbf{Action Regions} Spatial localization of actions in single frames for network interpretation}
\end{subfigure}
\begin{subfigure}{0.8\linewidth}
  \centering
  \includegraphics[width=\linewidth]{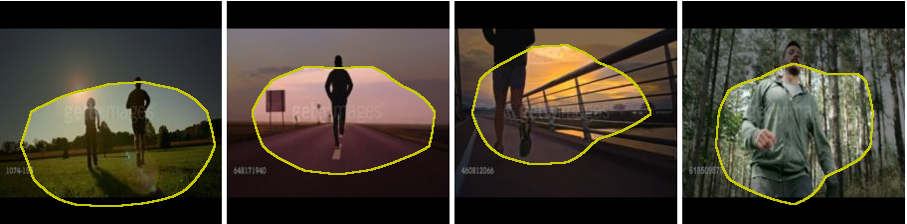}
  \caption{\textbf{Action Concepts} Interpretable action features learned by a trained model (i.e. jogging)}
\end{subfigure}
\end{figure}

In this paper we present the Multi-Moments in Time dataset (M-MiT).  This is a multi-label extension to the Moments in Time dataset (MiT) \cite{monfortmoments} which includes one million 3-second videos each with a human annotated action label.

Videos by their nature are dynamic.  In contrast to images, events can evolve over time and a single action label for an event may not fully capture the set of actions being depicted.  For example, a short video of a person \emph{raising} their hand and \emph{snapping} their fingers before \emph{laughing} has multiple actions (e.g. \emph{raising}, \emph{snapping}, \emph{laughing}) as part of the main activity.  Single label action datasets for video so do not provide annotations on this full set of information and instead label a single action in each video.  This leads to information loss in training, as only partial labels are provided per video and results in an incomplete/incorrect evaluation of trained models.  For example, an action model may return \emph{laughing} for the video previously described but the single label provided by the dataset for the video may be \emph{snapping}.  In this case evaluation will flag this as an incorrect prediction when it is actually correctly identifying a present action. This problem is exacerbated when we consider simultaneous actions that may take place in other parts of the video (e.g. a person \emph{coughing} in the background).  See Section \ref{sec:single_to_multi} for empirical validation of these claims.

Several large-scale video datasets provide a large diversity and coverage in terms of the categories of activities and exemplars they capture \cite{kay2017kinetics,goyal2017something,monfortmoments}.  However, these labeled datasets only provide a single annotated label for each video and this label may not cover the rich spectrum of events occurring in the video.  For example a video of an audience \emph{applauding} may also include a person on a stage \emph{performing}, \emph{playing music}, \emph{singing}, \emph{dancing}, etc.  There are a number of existing multi-label datasets for action detection in video \cite{yeung2015every,gu2017ava,DBLP:journals/corr/SigurdssonVWFLG16,Damen2018EPICKITCHENS}, however these datasets either do not share the scale in diversity of content and the number of annotated videos of the large single label datasets, like Kinetics \cite{kay2017kinetics} or Moments in Time \cite{monfortmoments}, or suffer from low specificity in their label set, as in Youtube-8M \cite{youtube8m} which contains a large set of videos with multiple weak labels sourced from YouTube topic tags.  A likely cause is the increased cost of collecting multiple labels for for a large set of videos compared to single label annotations.  However, these large-scale datasets have been shown to be important for training models that can capture robust representations that can be used to transfer to smaller datasets downstream \cite{kay2017kinetics,monfortmoments}.  With this in mind we have decided to build a large multi-label dataset that captures the scale and diversity of large single label video datasets.  While the visual, audio and semantic complexity of each video is challenging to fully annotate, we took a step toward this goal by extending the Moments in Time dataset, to contain multiple action labels describing one or more events occurring in each clip.

Building a multi-label dataset introduces new challenges in how to train and analyze models for multi-label action detection, including loss function to optimize model learning such as to take advantage of distinct labels for the same visual inputs.  For model's representation analysis, recently introduced methods such as Class Activation Mapping (CAM) \cite{zhou2016learning} and Network Dissection (NetDissect) \cite{netdissect2017} focused on single label interpretation (CAM) or did not provide methods for analyzing learned action concepts (NetDissect).  We address these limitations by extending both of these approaches to multi-action models (see Figure 1).

In Section \ref{sec:cam} we present a multi-label extension to Class Activation Mapping that identifies the important image regions for predicting multiple simultaneous actions in a given scene. In Section \ref{sec:netdissect} we outline our approach for adding action concepts to the NetDissect framework for interpreting internal units of a deep network. We additionally examine different multi-label loss functions applied to our dataset in Section \ref{sec:results}.  This includes a modification to the recently introduced LSEP loss function \cite{Li2017ImprovingPR} to support weighted learning for imbalanced datasets to handle the natural long tail distribution of actions in video. In the next section, we describe related work in the area followed by a description of the Multi-Moments in Time dataset and our annotation procedure. Overall the key contributions of this paper include:

\begin{itemize}
    \item \textbf{Multi-Moments in Time (M-MiT)}: A large-scale multi-label action dataset for video understanding with over two million action labels
\footnote{The data is available on our site, \url{http://moments.csail.mit.edu}.}.
    \item \textbf{wLSEP}: A novel multi-label loss function that supports learning from an imbalanced class distribution where some classes have more examples than others.
    \item \textbf{mCAM}: Multi-Label Class Activation Mapping for identifying multiple important visual features for model predictions.
    \item \textbf{Action Network Dissection}: We present a single frame dataset (Action Boxes) with bounding boxes on visible actions that we use for incorporating \textbf{action concepts} into Network Dissection \cite{netdissect2017} to identify key interpretable features learned by action models.
\end{itemize}

\section{Related work}

\subsection{Action Datasets for Video Understanding}
Video understanding has recently seen fast progress partly due to the availability of large scale video datasets for action recognition such as Kinetics \cite{kay2017kinetics},  Moments in Time \cite{monfortmoments} and ActivityNet \cite{caba2015activitynet}.  These large datasets are used to pretrain large video models that can be fine-tuned on smaller action recognition datasets such as UCF101 \cite{soomro2012ucf101}, HMDB \cite{10.1007/978-3-642-33374-3_41},  THUMOS \cite{jiang2014thumos}, and ``something something''  \cite{goyal2017something}.  Fine-tuning in this way allows the models to learn robust representations from the large variety of videos in the large datasets and transfer their features to the smaller sets.

Each of these datasets contains a single action label for each video clip provided.  This has been shown to be a limitation as actions tend to be intricately connected \cite{yeung2015every} and a single label on a video often misses background events and richer descriptions of events.  For example, a video of a person swimming in a pool may be labeled with the action "swimming".  This is correct, but consider that they are participating in a swim meet and are racing other swimmers.  Adding the labels "competing" and "racing" may help capture this information.  Additionally, there may be a person "running" along side the pool in the background.  Our proposed dataset aims to address this problem and provide multiple labels for each video.  There have been a number of multi-label video datasets introduced for action detection.  However, they either tend to be much smaller in scale than the large single-label datasets, such as MultiThumos \cite{yeung2015every}, AVA \cite{gu2017ava} and Charades \cite{DBLP:journals/corr/SigurdssonVWFLG16}, or constrained to specific domains, such as EPIC-KITCHENS \cite{Damen2018EPICKITCHENS}.  Our proposed Multi-Moments in Time dataset is large-scale, diverse and includes manually annotated action labels for each video.

\subsection{Models for Video Understanding}

To take advantage of the different datasets described in the previous section, many different models have been proposed.  A popular architecture is the two-stream CNNs \cite{simonyan2014two}
which separately processes optical flow and RGB frames.  3D CNNs \cite{tran2015learning} use a 3-dimensional kernel to learn temporal information directly from the frame sequence of a video and 
I3D models \cite{carreira2017quo} combine 3D CNNs with optical flow to form a two-stream 3D network ``inflated'' from 2D filters pre-trained on ImageNet \cite{deng2009imagenet}. More recently a temporal shift module (TSM) has been used to integrate temporal information into 2D models by shifting frame representations across the temporal dimension \cite{Lin_2019_ICCV}.  We compare results using both I3D and TSM baseline models on our proposed dataset in Section \ref{sec:results}.

\subsection{Multi-Label Optimization}

Multi-label optimization is a common problem in object dectection where multiple objects may be present in a single image. Given the variety of appearances of objects within a scene, it is challenging for global CNN features to correctly predict multiple labels. Approaches instead use object proposals \cite{gong2013deep,yang2016exploit} or learn spatial co-occurrences of labels using LSTMs \cite{wang2016cnn,zhang2018multi}. Convolutional neural networks (CNN) have also been applied on raw images of multiple objects to learn image-level deep visual representations for multi-label classification \cite{wang2016cnn}.

To improve performance on multi-label detection tasks, different loss functions have been proposed.  A common approach is binary cross entropy which optimizes each class label individually. 
However, when treating each class individually, it is also difficult to learn the correlations between different classes \cite{Elisseeff:2001:KMM:2980539.2980628, Zhang:2006:MNN:1159162.1159294,Li2017ImprovingPR}. Additionally, this approach may incorrectly penalize some examples which do not have full label coverage as it assumes the absence of a label is a negative label.  Another approach is pair-wise ranking \cite{Weston:2011:WSU:2283696.2283856} which encourages the model to generally assign higher ranks to positive labels.  This method has the added benefit of reducing the strength of a models' mistake as incorrect predictions tend to still include highly ranked positive labels.  WARP \cite{gong2013deep,Weston:2011:WSU:2283696.2283856} expands on this by including a monotonically increasing weighting function that increases the error for positive labels that are poorly ranked, thus prioritizing them in learning.  BP-MLL \cite{Zhang:2006:MNN:1159162.1159294} is another approach that provides a smooth calculation of the ranking error but suffers from exceedingly large values when the positive classes are poorly ranked and the vocabulary size is large.  LSEP \cite{Li2017ImprovingPR} is a variation of the BP-MLL loss function which addresses it's numerical stability issues but can become dominated by poorly ranked positive classes and does not allow for weighting the loss accross classes in imbalanced datasets.  We propose a modification of the LSEP loss function to reduce the effect of these poorly ranked positive classes while integrating a class weighting term that aims to balance learning in datasets with imbalanced label distributions which are common in multi-label problems.

\subsection{Model Interpretation}

Recent work in network interpretation performs a weighted inflation of the feature maps in the final convolutional layer of a CNN to visualize the important visual regions the network is using to make a prediction \cite{zhou2015cnnlocalization}.  These class activation maps (CAMs) can be thought of as a method for visualizing the learned attention model of the network and have been shown to localize actions occurring in videos \cite{monfortmoments}.  Additionally, network interpretation has been extended to not just visualize important visual regions for a prediction but to also identify the different concepts a network has learned \cite{Bau_2017_CVPR,8417924}.  This is important for understanding the representation of the network and diagnosing class biases that the network is learning.  However, the previous work in network dissection (NetDissect) does not include action concepts in its interpretation instead relying on objects, scenes, object parts, textures and materials.  In this paper we extend the dataset used (Broden dataset) for this interpretation to include actions as well so that we can better interpret action recognition models.  We also extend class activation mapping to visualize specific regions important to different actions in images with multiple detected actions.

\section{Building a Multi-Action Dataset}

In this section we describe our approach to building and annotating our Multi-Moments in Time dataset (M-MiT) and present summary statistics on our dataset characteristics.

%Our goal with this project is to extend the existing Moments in Time dataset to include multiple action labels per video to take another step toward improving our understanding of the diverse and dynamic events that take place in short videos.  With this multi-label dataset we will be able to better evaluate our existing models and train new models that can better leverage the amount of information in each video as well as the relationships between different actions.

\subsection{Annotation}

We began by annotating our added action labels using the same process as the Moments in Time dataset. The annotation phase used Amazon Mechanical Turk for crowd sourcing where each worker is presented with a video-verb pair and asked to respond Yes or No if the action is either seen or heard in the video.  We additionally embed ground truth video-action pairs into every set that have been manually verified by our team to either be a positive pair (the action is in the video) or a negative pair (the action is not in the video).  We randomly replace 10\% of the questions in every annotation set with these ground truth pairs and use them to evaluate workers.  If a worker gets more than one of these questions wrong then we do not let them submit the set and if a worker repeatedly submits sets with an incorrect ground truth response then we flag the worker and prevent them from completing future jobs.  Each action-video pair is annotated at least three times for the training set and at least 5 times for the validation set.  This ensures high-quality annotations are collected efficiently.  We refer the reader to the Moments in Time paper for more details \cite{monfortmoments}.

\subsubsection{Generating Action Candidates}
 A difficulty of the annotation task is to choose candidate actions that are likely to return positive responses (i.e. actions that occur in the videos).   We generated candidate actions for each video using few different techniques.
 
 \noindent \textbf{Approach 1:} First, we began by selecting candidates using WordNet \cite{Miller:1995:WLD:219717.219748} relationships where we iteratively picked actions for annotation that were closely linked in the WordNet semantic graph to the existing action labels for each video.  We constrained ourselves to the original Moments in Time vocabulary of 339 actions in order to simplify this process and stopped our candidate selection when the harvest rate (positive worker response rate) dropped below 20\%.  For example, if we have a video with the action \emph{running}, the WordNet graph structure shows a short semantic path (distance between word nodes in the graph) to the action \emph{exercising} and begin annotating all videos that have the action \emph{running} with the action \emph{exercising}.  We iteratively increase the node distance used to find related actions in the graph until the harvest drops below the 20\%.  It is important to note that the hierarchy in action relationships are less well defined than that of objects.  For example, not every video of \emph{eating} is defined by the action \emph{dining}.  Similarly there are videos in the dataset of animals \emph{running} that are \emph{chasing} each other which do not relate to \emph{exercising} in the same way a video of a person \emph{jogging} does.  It is important to note that these approaches are to generate action candidates that we verify through multiple rounds of human annotation.  In this way we are able to verify whether a video of a person \emph{running} is in fact \emph{jogging} as opposed to \emph{sprinting} and/or \emph{chasing}.
 
  \noindent \textbf{Approach 2:} Similar to the WordNet approach, we use Word2Vec \cite{DBLP:journals/corr/abs-1301-3781} similarity scores to select action candidates based on the existing action labels for a video.  Word2Vec allows us to generate candidates for actions that describe commonly co-occurring events such as \emph{stirring} and \emph{boiling} or \emph{running} and \emph{jumping}.  Similarly, this approach helps to generate opposite, but commonly co-ocurring action pairs such as \emph{opening} and \emph{closing} or \emph{throwing} and \emph{catching}.  A limitation of this approach is that it is based on NLP embeddings and may not be fully reflective of action relationships in video.
 
\begin{figure}
\centering
\resizebox {0.9\columnwidth} {!} {
 \begin{tikzpicture}[node distance = 1cm, thick]% 
        \node (1) {running};
        \node (2) [right=of 1] {competing};
        \node (3) [above=of 2] {jumping};
        \node (4) [below=of 2] {cheering};
        \node (5) [right=of 4] {};
        \node (7) [right=of 4] {shouting};
        \node (6) [above=of 7] {applauding};
        \node (8) [right=of 3] {falling};
        \node (9) [right=of 8] {rolling};
        \node (10) [right=of 6] {clapping};
        \draw[->] (1) -- node [midway,above] {} (3);
        \draw[->] (1) -- node [midway,above] {} (4);
        \draw[->] (4) -- node [midway,above] {} (6);
        \draw[->] (4) -- node [midway,above] {} (7);
        \draw[->] (3) -- node [midway,above] {} (8);
        \draw[->] (8) -- node [midway,above] {} (9);
        \draw[->] (1) -- node [midway,above] {} (2);
        \draw[->] (6) -- node [midway,above] {} (10);
\end{tikzpicture}
}
\caption{An example of the path of generating new candidate verbs from previously annotated classes.}
\label{fig:candidate_generation}
\end{figure}
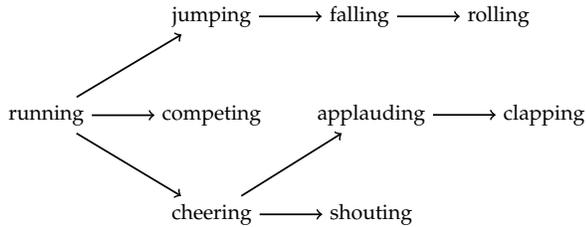
 
  \noindent \textbf{Approach 3:} The final approach we use for candidate selection is to run a trained model over the videos and select the top-1 predictions for each video that are not currently annotated. The model used for this task was trained on the single label dataset and is provided by the authors of the original work \cite{monfortmoments}. We restricted the actions to the top-1 predictions as taking a larger range of predictions quickly dropped the harvest rate which limited our ability to scale. This approach can capture events that may be seemingly unrelated but still co-occurring in a video such as a child \emph{jumping} in a doorway while an adult is \emph{cooking} in a kitchen.  This does introduce some model bias into the candidate generation process but when combined with the other approaches we consider it acceptable to reach a stronger coverage of the events in our videos.
 
 For each of the above approaches we regenerate new candidates as new labels are verified through annotation.  Figure \ref{fig:candidate_generation} outlines the candidate generation path of a video that was originally only labeled with the action \emph{running}.  In this case we first generate and annotate the action candidates \emph{jumping}, \emph{competing} and \emph{cheering} which then eventually lead to subsequent annotated actions such as \emph{clapping} and \emph{rolling} such that we are able to provide a much more thorough description of the actions taking place in the video than simply \emph{running}.  Also note that \emph{cheering}, \emph{applauding}, \emph{clapping} and \emph{shouting} are auditory actions that may not be visible in the video itself but instead heard.
 
 These methods of candidate generation allow us to efficiently annotate multiple actions in each video, but it does result in videos where we do not annotate every present action.  We consider this an acceptable trade-off for the ability to scale to such a large set of labels and we ensure that in training models we do not directly penalize predictions for actions that are not labeled (see Section \ref{sec:loss}).
 
 \begin{figure}
  \centering
  \includegraphics[width=0.9\linewidth]{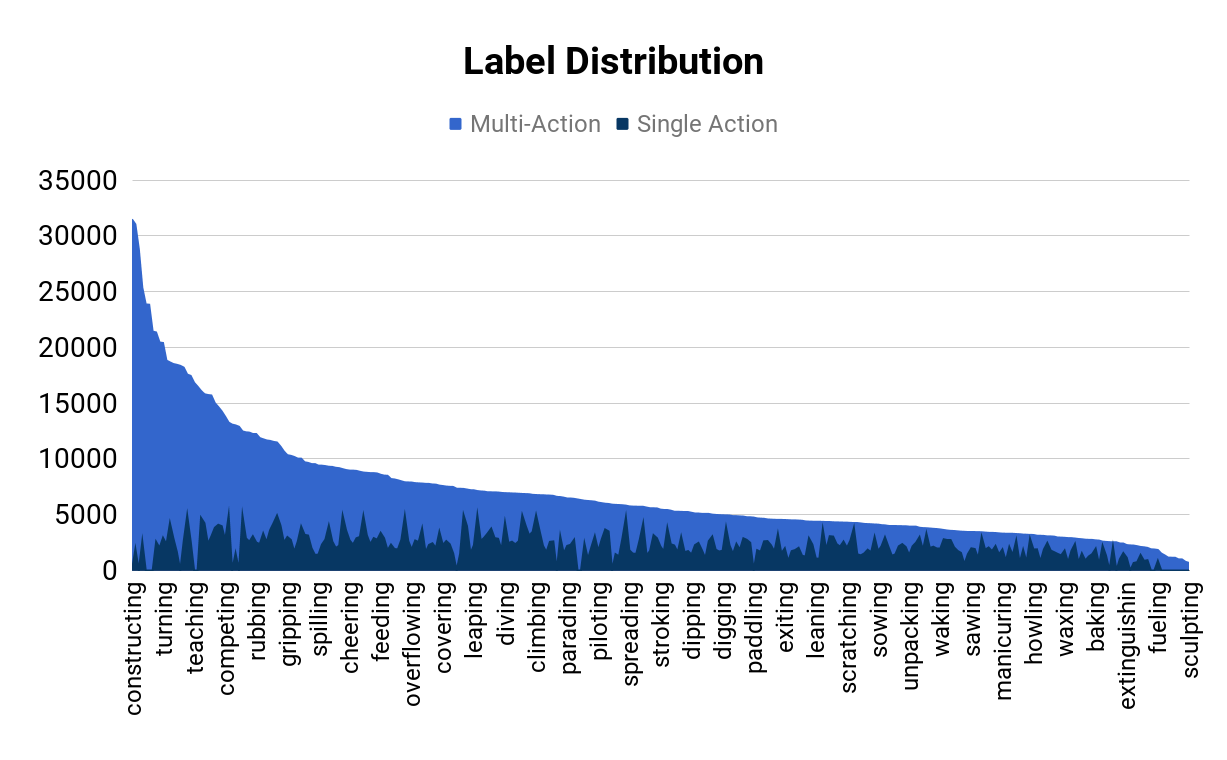}
  \caption{The label distribution of our proposed Multi-Moments in Time dataset compared to the Moments in Time dataset}
  \label{fig:distribution}
\end{figure}

\subsection{Dataset Statistics}

There are 802,244 video-label pairs in the training set, and 33,900 in the validation set of the Moments in Time Dataset.  We increased this dataset to 2.01 million labels for 1.02 million videos by adding new videos, generating and annotating action candidates as described in the previous section and adding new action classes to the dataset.  

Our \emph{Multi-Moments in Time} dataset includes 292 annotated action classes where new actions have been added to the Moments in Time vocabulary (e.g. \emph{skateboarding}), ambiguous/noisy actions were removed (e.g. \emph{working}) and similar actions have been merged into a single class (e.g. \emph{rising} and \emph{ascending} -> \emph{ascending/rising}). We additionally added novel actions not found in the original Moments in Time (e.g. \emph{unplugging}) and removed actions that we deemed either too vague or noisy based on a random sample of 500 videos per class (e.g. \emph{working}).  Figure \ref{fig:label_changes} shows the changes made to the class vocabulary from MiT.

\begin{figure}
\small
\begin{framed}
\centering
\textbf{Combined}\\
\begin{framed}
\begin{minipage}{0.48\linewidth}
\footnotesize
stroking/petting\\
smiling/grinning\\
stacking/piling\\
breaking/destroying\\
filming/photographing\\
cleaning/washing\\
constructing/assembling\\
ascending/rising\\
combusting/burning\\
descending/lowering\\
\end{minipage}
\begin{minipage}{0.48\linewidth}
\footnotesize
teaching/instructing\\
rotating/spinning\\
shoveling/digging\\
eating/feeding\\
tearing/ripping\\
smelling/sniffing\\
hitting/colliding\\
laughing/giggling\\
buying/selling/shopping\\
snuggling/cuddling/hugging\\
\end{minipage}
\end{framed}
\textbf{Added}\\
\begin{framed}
\begin{minipage}{0.24\linewidth}
\footnotesize
paddling\\
weightlifting\\
parading\\
squeezing\\
sculpting\\
punting\\
texting\\
\end{minipage}
\begin{minipage}{0.24\linewidth}
\footnotesize
waxing\\
scooping\\
erasing\\
laying\\
bandaging\\
shivering\\
dialing\\
\end{minipage}
\begin{minipage}{0.24\linewidth}
\footnotesize
inserting\\
unplugging\\
rubbing\\
massaging\\
kayaking\\
sleeping\\
flossing\\
\end{minipage}
\begin{minipage}{0.24\linewidth}
\footnotesize
ironing\\
ice+skating\\
approaching\\
snowboarding\\
skateboarding\\
smashing\\
gasping\\
\end{minipage}
\end{framed}
\textbf{Removed}\\
\begin{framed}
\begin{minipage}{0.24\linewidth}
\footnotesize
coaching\\
boarding\\
flicking\\
stopping\\
working\\
asking\\
starting\\
\end{minipage}
\begin{minipage}{0.24\linewidth}
\footnotesize
serving\\
sketching\\
paying\\
putting\\
talking\\
entering\\
placing\\
\end{minipage}
\begin{minipage}{0.24\linewidth}
\footnotesize
exiting\\
blocking\\
cramming\\
tickling\\
joining\\
raising\\
watering\\
\end{minipage}
\begin{minipage}{0.24\linewidth}
\footnotesize
imitating\\
drenching\\
bouncing\\
guarding\\
building\\
leaning\\
playing\\
\end{minipage}
\end{framed}
\end{framed}
\caption{Lists of vocabulary differences in M-MiT compared to MiT.  We combined similar classes, removed ambiguous/noisy classes and added some new classes found during annotation.}
\label{fig:label_changes}
\end{figure}

This new vocabulary should cover an increased breadth of events while improving the boundary between different classes.  Using this new action set we were able to increase the training set to include over 2 million labels where 553,535 videos are annotated with more than one label and 257,491 videos are annotated with three or more labels.  In addition, we have created new validation and test sets each consisting of 10K videos with over 30K labels each.  Figure \ref{fig:distribution} shows a comparison of the distribution of our proposed M-MiT dataset to single label MiT dataset.  The classes in the training set have an average of 6,432 example videos and a median of 5,478 videos while the classes in the validation set have an average of 96 example videos and a median of 31 videos.

\section{Multi-Label Loss Functions}
\label{sec:loss}

In this section we present a set of multi-label loss functions that we use to train models on our multi-action dataset.   
For each loss we normalize by the number of labels in each data example to handle the variability in the number of labels per video and incorporate a class weighting term $w_i$ that helps to balance learning when the training set has an imbalanced number of examples per label.  This imbalance is common for action datasets as action labels tend to follow a long tail distribution in practice.  Additionally, in prior work sampling was used to address the quadratic complexity of the different loss functions \cite{Li2017ImprovingPR}.  However we have found that parallelizing the loss computation through matrix operations eliminates the need for sampling.  We compare the results of each approach in Section \ref{sec:results}.

\subsection{BCE}

A common approach to multi-label optimization is to optimize for binary cross entropy and treat each label as an independent classifier.  Given the output of a model, $x_i$ for class $i$ and an indicator of whether this class is positive, $y_i$,
\begin{align}
    \mathcal{L}_{BCE}=-w_i[y_i \log x_i + (1-y_i)\log(1-x_i)],
    \label{eq:bce}
\end{align}
where $w_i$ is a weight balancing term that scales the strength of the loss applied to class $i$ to improve optimization for underrepresented classes in imbalanced datasets.

However, this has been shown to not consider correlations between different classes \cite{Elisseeff:2001:KMM:2980539.2980628, Zhang:2006:MNN:1159162.1159294,Li2017ImprovingPR} and makes the assumption that cases where a class does not have a positive label are negative examples.  While we have verified all positive labels in the proposed dataset, we do not assume that an unlabeled class is guaranteed to not be present.  

\subsection{WARP}
Pair-wise ranking presents another approach to multi-label optimization stemming from the motivation that while it is important to correctly classify a positive label, it is also important to reduce the strength of a mistake by encouraging the model to assign higher ranks to positive labels \cite{Weston:2011:WSU:2283696.2283856}.  The WARP loss function \cite{gong2013deep,Weston:2011:WSU:2283696.2283856} proposes a weighted pair-wise ranking function that prioritizes poorly ranked positive classes via an additional monotonically increasing weighting function $\mathcal{W}(x_i)$,
\begin{align}
    \mathcal{L}_{WARP}=\frac{1}{|\mathcal{Y}|}\sum_{i\in\mathcal{Y}} w_i \mathcal{W}(\mathcal{R}(x_i))\sum_{j\notin\mathcal{Y}} \max(0,1+x_j-x_i),
    \label{eq:warp}
\end{align}
where $\mathcal{Y}$ represents the set of positively labeled classes.

  The rank of a predicted class, $\mathcal{R}(x)$, is used to increase the error penalization for lower ranked classes.  In practice we set $\mathcal{W}(r)=\sum_{k=1}^{r}\frac{1}{k}$.  %The limitation of WARP lies in the non-smooth nature of the loss function which can make it difficult to optimize.

\begin{figure*}
\centering
\begin{minipage}{0.8\linewidth}
\centering
\includegraphics[width=\linewidth]{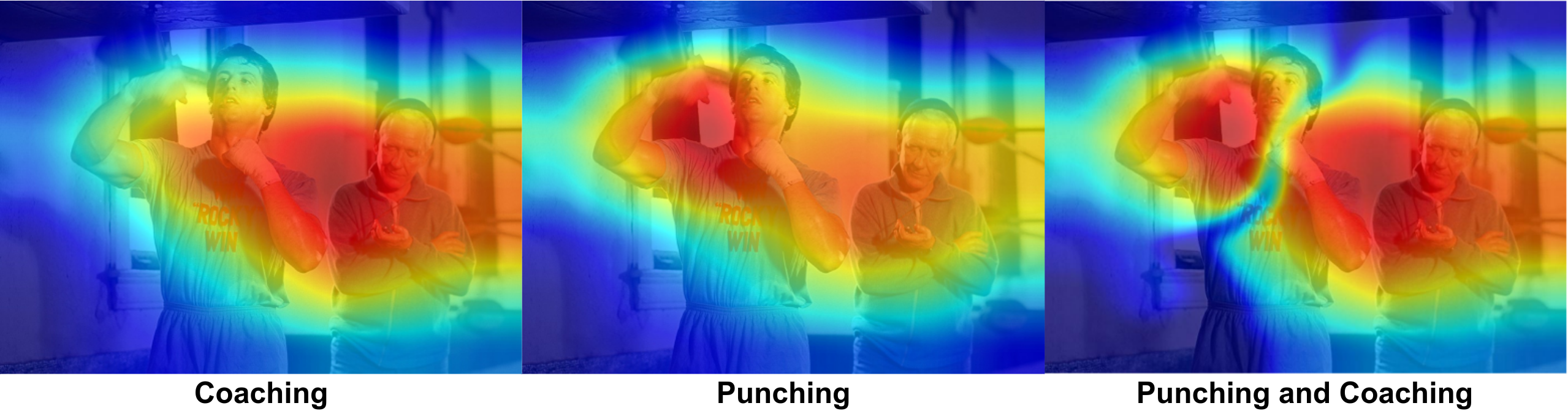}
\caption{\textbf{Region Separation:} Example showing single class CAM images and the separation of relevant features in a multi-class CAM image.  The red regions specify important areas of the image used by the model to infer the detected action.}
\label{fig:multi_cam}
\end{minipage}
\begin{minipage}{0.8\linewidth}
\centering
\includegraphics[width=\linewidth]{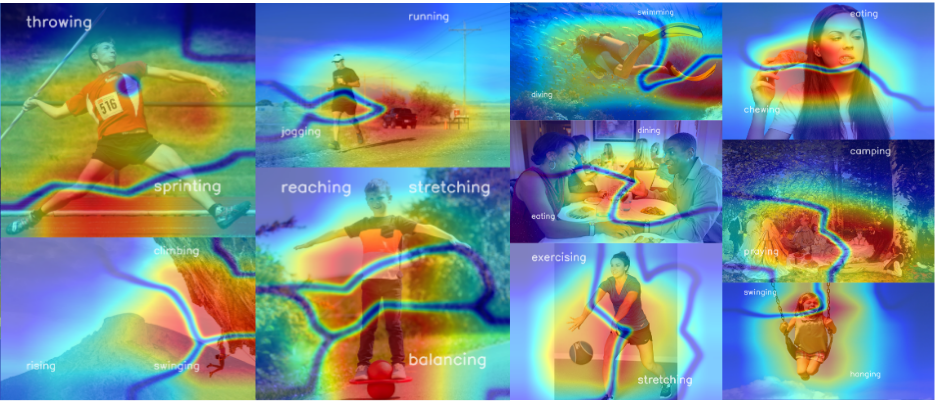}
\caption{\textbf{Multi-CAM Examples:} Multi-class CAM images for a variety of scenes with simultaneous actions.  Action labels are placed near the important image regions used by the model for identifying each specific action.  This area is signified by the red overlay with blue edges separating image regions distinct to each detected action showing that our model is able to localize multiple actions present in each scene despite not being trained for localization.}
\label{fig:multi_cam_collage}
\end{minipage}
\end{figure*}

\subsection{LSEP}

The recently proposed multi-label ranking function LSEP \cite{Li2017ImprovingPR} takes the log of the BP-MLL function \cite{Zhang:2006:MNN:1159162.1159294}, with the addition of a single bias term, to increase the numerical stability,
\begin{align}
    \mathcal{L}_{LSEP}=\log\big(1+\sum_{i\in\mathcal{Y}}\sum_{j\notin\mathcal{Y}} e^{x_j-x_i}\big).
    \label{eq:lsep}
\end{align}
This prioritizes positive classes that are poorly ranked without the need of an additional weighting function as in WARP.  However, poorly ranked classes can dominate the loss and the function does not allow for a weighting term to be included to improve optimization in imbalanced datasets.

\subsection*{wLSEP}

We propose modifying the LSEP loss to add a weight balancing term to aid optimization in our imbalanced dataset.  This is not straight forward as the gradient computation for each class is non-separable from other classes.  We address this by modifying the LSEP loss to apply individually to each class with a positive label allowing us to simply add a weight term, $w_i$, to the function,
\begin{align}
    \mathcal{L}_{wLSEP}=\frac{1}{|\mathcal{Y}|}\sum_{i\in\mathcal{Y}}w_i \log\big(1+\sum_{j\notin\mathcal{Y}} e^{x_j-x_i}\big).
    \label{eq:wlsep}
\end{align}
This looses the prioritization of low ranked positive classes in the original loss function by summing the softmax of the ranking error ($x_j-x_i$) for each positive label $i$, however it avoids the problem of taking a global softmax over all positive labels which prevents the loss from being dominated by poorly ranked positive labels.

\section{Analyzing Multi-Label Models}
\subsection{Multi-Label Class Activation Mapping}
\label{sec:cam}

Here we extend the class activation mapping (CAM) \cite{zhou2015cnnlocalization} technique to multi-label tasks. The simplest approach is to inflate the CAM for each predicted action, taking the maximum value of each map so that we can visualize the discriminative features used for our multi-label prediction.  However, in practice, the combined CAM filters cover a large area of the input image making it difficult to gain a useful understanding of the models' decisions.  CAM is very useful for identifying the key visual features that contribute to a model's prediction.  In our multi-label setting we want to identify the unique features that are contributing to each prediction.  This is clear for very different actions such as \emph{jumping} and \emph{swimming} but the distinct features used to detect \emph{dining} and \emph{eating} may be more ambiguous.  We care about separating these regions to better understand our model and how it is making it's decisions.  This can help us identify any biases learned by the model and may help us understand small differences between seemingly similar actions.

To address this issue, before we take the maximum value of each actions CAM, we first compare the different CAM filters for each of our predicted classes and when two filters with a cosine distance greater than $1^{-4}$ have similar values at the same pixel locations we set the values of the pixels for both filters to zero.  This eliminates the overlapping area and, after performing maximum pooling over the altered CAM filters, creates a boundary between the distinct features associated with each predicted action.  Only altering filters with a cosine distance greater than some threshold, in this case $1^{-4}$, ensures that detections that use the same regions do not have their CAMs erased. In practice we found that this value strikes a strong balance in generating distinct visualizations of regions used by the model for multiple detections while merging highly similar regions used by the model such as \emph{emptying} and \emph{pouring} in the image of wine being poured into a glass in Figure \ref{fig:multi_cam_collage}.  If a strict comparison of features used between two detected actions in an image is desired then this threshold should be set to zero.  To improve the visualizations we also apply Gaussian smoothing with a 5x5 kernel to reduce the sharpness of some of the boundary edges.  We show the results of this method for analyzing multi-label predictions in Section \ref{sec:multiCamResults}.

\subsection{Identifying Learned Action Features}
\label{sec:netdissect} 

To better analyze our learned action models we extend the set of concepts used by NetDissect \cite{Bau_2017_CVPR,8417924} to include actions.  NetDissect uses the Broden dataset, an image segmentation dataset consisting of pixel-level annotations of objects, scenes, parts, materials and colors.  For each image input into a model the labeled segmentations are compared to the different activation regions in internal feature maps of the model.  For example, when the activation regions for a feature map share a strong correlation, quantified using IoU, with the segmented regions of a specific object class in a set of images then the feature is interpreted to be correlated to that object.  We refer to the NetDissect paper for more information.  Here, we extend the Broden dataset to include images with segmented action regions so that we can correlate features with action classes.

\subsubsection{Annotation}

As with annotating the action labels in the videos, we collect bounding box annotations via Amazon Mechanical Turk (AMT).  We begin by selecting a single frame from the center of 500 randomly selected videos from each of the action classes in Moments in Time \cite{monfortmoments}, Kinetics \cite{kay2017kinetics} and the Something-Something \cite{goyal2017something}  datasets.  We then present a binary annotation task to the workers on AMT asking if an action from the source video's label set is visible in the frame shown.  This binary interface is very similar to that used for collecting the action labels for the Moments in Time dataset \cite{monfortmoments} with the main difference being the use of images rather than video.  We run this task for at least 2 rounds of annotation to verify that the action is visible in each frame.  We then take the set of verified action-frame pairs and pass them to a separate annotation interface on AMT that asks the workers to select the regions most important in the image for identifying the action.  Multiple regions can be selected for an image, as in the jogging example in Figure \ref{fig:action_regions}, and the workers are allowed to skip an image if there are no useful regions for detecting the action (i.e. the action is not visible in the image).

We run this region selection task through multiple rounds and only consider overlapping regions from the different rounds as most important for detecting the actions.  After this stage the regions selected are cropped from the original images and passed through the binary annotation task previously described for a final verification that the actions are present and recognizable in the selected regions.  After our complete annotation process our total set of verified images with segmented action regions consists of 67,468 images from 549 different action classes.  Figure \ref{fig:action_regions} displays some examples of the selected regions collected through this process.

\subsubsection{Action Interpretation}

To integrate our new action region dataset into the Network Dissection framework we first consider each selected region to be a mask on the segmented area of the image relating to the action.  This is similar to part, material and object masks used for other segmentation datasets \cite{zhou2016semantic,chen_cvpr14,mottaghi_cvpr14,bell13opensurfaces}.  With the data formatted in this manner we extend the Broden dataset to include our action segmentations and extract the set of learned action concepts detected via NetDissect.  This process allows us to identify not just object, scene, texture and color concepts learned by our models, but action concepts as well.  In Section \ref{sec:interpretationResults} we show some of the key results from interpreting action networks in this way.

\section{Baseline Results}
\label{sec:results}

We trained two architectures on our dataset, an inflated 3D ResNet-50 (I3D) \cite{carreira2017quo} for the visual modality and a SoundNet network \cite{NIPS2016_6146} for learning audio features and compare across different multi-label loss functions.

\subsection{Models}

\begin{table}[tb]
    \centering
    \footnotesize
    \begin{tabular}{  c | c | c | c | c | c}
        %\hline
        %\toprule
        \multirow{2}{*}{\textbf{Model}} & \multirow{2}{*}{\textbf{Loss}} & \multirow{2}{*}{\textbf{Top-1}} & \multirow{2}{*}{\textbf{Top-5}} & \textbf{Micro} & \textbf{Macro} \\
        & & & & \textbf{mAP} & \textbf{mAP}\\
        \hline
        \multirow{4}{*}{\textbf{I3D}}
        & \textbf{BCE} & 52.9 & 77.8 & 55.4 & 37.8\\
        & \textbf{WARP} & 51.9 & 79.2 & 56.5 & 32.7 \\
        & \textbf{LSEP} & 52.7 & 78.8 & 56.1 & 31.7\\
        & \textbf{wLSEP} & \textbf{58.5} & \textbf{81.4} & \textbf{61.7} & \textbf{39.4}\\
        \hline
        \multirow{4}{*}{\textbf{TSM}}
        & \textbf{BCE} & 56.9 & 80.1 & 58.7 & \textbf{37.8} \\
        & \textbf{WARP} & 55.8 & 81.0 & 60.0 & 33.4 \\
        & \textbf{LSEP} & 55.2 & 81.4 & 59.8 & 27.6 \\
        & \textbf{wLSEP} & \textbf{58.1} & \textbf{81.6} & \textbf{62.4} & 35.4 \\
        \hline
        \multirow{4}{*}{\textbf{Audio}}
        & \textbf{BCE} & 6.8 & 19.3 & 6.9 & 3.1\\
        & \textbf{WARP} & 6.3 & 18.6 & 6.6 & 2.8\\
        & \textbf{LSEP} & 5.3 & 16.1 & 6.0 & 2.6\\
        & \textbf{wLSEP} & \textbf{6.9} & \textbf{20.9} & \textbf{6.9} & \textbf{3.2} \\
        %\hline
        \hline
        \textbf{Fusion} & \textbf{wLSEP} & \textbf{59.3} & \textbf{82.8} & \textbf{61.8} & \textbf{41.1} \\
        \hline
    \end{tabular}
    \vspace{10pt}
    \caption{\textbf{Validation Results:} Performance of the baseline models with different loss functions on the multi-label validation set. We show Top1, Top5 and both micro and macro mAP.}
    \label{table:results}
\end{table}

\begin{table}[tb]
    \centering
    \footnotesize
    \begin{tabular}{  c | c | c | c | c | c}
        \textbf{Training} & \textbf{Evaluation} & \multirow{2}{*}{\textbf{Top-1}} & \multirow{2}{*}{\textbf{Top-5}} & \textbf{Micro} & \textbf{Macro} \\
        \textbf{Dataset} &\textbf{Dataset} & & & \textbf{mAP} & \textbf{mAP}\\
        \hline
        \textbf{MiT} & \textbf{MiT-S} & 23.0 & 53.0 & 36.7 & 18.9\\
        \textbf{MiT} & \textbf{M-MiT-S} & 40.0 & 68.5 & 42.1 & 24.5\\
        \textbf{M-MiT} & \textbf{M-MiT-S} & 45.0 & 79.1 & 54.5 & 32.7\\
        \hline
    \end{tabular}
    \vspace{10pt}
    \caption{\textbf{MiT vs M-MiT Model Validation Results:} Performance of models trained on MiT and M-MiT on a subset of the M-MiT validation set containing only classes shared between MiT and M-MiT.  We evaluate with both a single label version (MiT-S) and a multi-label version (M-MiT-S). We show Top1, Top5 and both micro and macro mAP.  We see that multiple labels improves our evaluation of the MiT model while the M-MiT model achieves the best results.}
    \label{table:single_to_multi_results}
\end{table}

\begin{table}[tb]
    \centering
    \footnotesize
    \begin{tabular}{  c | c | c | c | c | c}
        %\hline
        %\toprule
        \multirow{2}{*}{\textbf{Dataset}} & \multirow{2}{*}{\textbf{Loss}} & \multirow{2}{*}{\textbf{Top-1}} & \multirow{2}{*}{\textbf{Top-5}} & \textbf{Micro} & \textbf{Macro} \\
        & & & & \textbf{mAP} & \textbf{mAP}\\
        \hline
        \multirow{4}{*}{\textbf{MS-COCO}}
        & \textbf{BCE} & 90.3 & 98.6 & 82.6 & 63.6\\
        & \textbf{WARP} & 92.0 & 98.8 & 83.7 & 60.9\\
        & \textbf{LSEP} & 94.0 & 98.8 & 85.4 & 61.0\\
        & \textbf{wLSEP} & \textbf{94.3} & \textbf{98.9} & \textbf{86.3} & \textbf{64.5}\\
        \hline
        \multirow{4}{*}{\textbf{VOC}}
        & \textbf{BCE} & 91.7 & 97.9 & 91.6 & 83.7\\
        & \textbf{WARP} & 93.2 & 99.4 & 92.9 & 85.1 \\
        & \textbf{LSEP} & 92.8 & 99.2 & 92.6 & 85.2\\
        & \textbf{wLSEP} & \textbf{93.8} & \textbf{99.5} & \textbf{93.4} & \textbf{87.2}\\
        \hline
        \multirow{4}{*}{\textbf{AVA}}
        & \textbf{BCE} & 58.3 & 92.1 & 65.5 & 9.3\\
        & \textbf{WARP} & 47.6 & 85.1 & 54.4 & 11.2 \\
        & \textbf{LSEP} & 71.0 & 95.7 & 71.2 & 10.9\\
        & \textbf{wLSEP} & \textbf{71.2} & \textbf{96.0} & \textbf{72.3} & \textbf{12.7}\\
        \hline
        \multirow{4}{*}{\textbf{MultiThumos}}
        & \textbf{BCE} & 63.4 & 91.1 & 65.5 & 58.9\\
        & \textbf{WARP} & 75.8 & 91.8 & 77.5 & 63.7\\
        & \textbf{LSEP} & 75.1 & 91.7 & 77.2 & 65.5\\
        & \textbf{wLSEP} & \textbf{75.9} & \textbf{92.0} & \textbf{77.6} & \textbf{68.0}\\
        \hline
    \end{tabular}
    \vspace{10pt}
    \caption{\textbf{Loss function comparison:} We validate our proposed wLSEP loss function on four different multi-label datasets.  A ResNet-50 model is trained for MS-COCO and VOC while a ResNet-50 I3D model is trained for AVA and MultiThumos}
    \label{table:other_dataset_results}
\end{table}

\subsubsection*{Inflated 3D Convolutional Networks (I3D)} I3D networks offer improved weight initialization by simply \emph{inflating} the convolutional and pooling kernels of pretrained 2D networks \cite{carreira2017quo}. This is done by initializing the inflated 3D kernel with pretrained weights from 2D models by repeating the parameters from the 2D kernel over the temporal dimension. This greatly improves learning efficiency and performance since 3D models contain a large number of parameters and are difficult to train from scratch.  For our experiments we use an inflated 3D ResNet-50 pretrained on ImageNet with 16 frames as input.

\subsubsection*{Temporal Shift Model (TSM)}
We additionally compare results using a ResNet-50 temporal shift module pretrained on ImageNet.  This model integrates temporal information into the 2D architecture by shifting frame representations across the temporal dimension \cite{Lin_2019_ICCV}.  In our experiments we used 8 frames as input into the model.

\subsubsection*{SoundNet Network (Audio)}
In MiT, action classes are labeled for both visual and auditory information.  Therefore we feel it would be incomplete to evaluate visual models and not include a model trained on audio. We finetune a SoundNet network \cite{NIPS2016_6146} which was pretrained on unlabeled videos from Flickr.

\subsubsection*{Spatio-temporal-Auditory Fusion}

We fuse the predictions from the audio and visual modalities by concatenating the spatio-temporal features of the I3D network with the auditory features from SoundNet and train a single linear layer to rank the detected action classes using the loss functions described in the following section.

\subsection{Performance Metrics}

\subsubsection*{Accuracy}

We report both the top-1 and top-5 classification accuracy for each of our models in order to be consistent with the results reported from the original Moments in Time paper \cite{monfortmoments}. Top-1 accuracy indicates the percentage of testing videos where the top predicted class is a positive label for the video.  Similarly, top-5 accuracy indicates the percentage of the testing videos where any of the top predicted 5 classes for a video is a positive label.

\subsubsection*{Mean Average Precision (mAP)}
We use mAP as our main evaluation metric as it captures errors in the ranking of relevant actions for a video. For each positive label, mAP computes the proportion of relevant labels ranked before it and averages over all of the labels.

In order to properly evaluate our models we report both the micro and the macro mAP.  The micro mAP is the mean average precision over all videos and the macro mAP is the average of the mAP for each class.  In the case of imbalanced datasets the micro mAP depicts the full performance of the model on the dataset while the macro mAP displays the models class-wise consistency.  These numbers can differ greatly as a high mAP in a highly represented class and a low mAP in all other classes can lead to a high micro mAP and a low macro mAP.

%\begin{table*}[tb]
%\centering
%\setlength{\tabcolsep}{2.5pt}
%\begin{tabular}{ c | c c c | c c c | c c c |}
%      %\toprule       
%       \multirow{2}{*}{\textbf{Loss}} & \multicolumn{3}{|c|}{\textbf{VOC}}  & \multicolumn{3}{|c|}{\textbf{NUS}} & %\multicolumn{3}{|c|}{\textbf{MS-COCO}}\\ 
%        & \textbf{Top-1}  & \textbf{Top-5} & \textbf{mAP} & \textbf{Top-1}  & \textbf{Top-5} & \textbf{mAP} & %\textbf{Top-1}  & \textbf{Top-5} & \textbf{mAP}  \\
%      \hline %\cmidrule{1-4}
%      \textbf{BCE}
%        & 0.953 & 0.995 & 0.945 & 0.338 & 0.782 & 0.387 & 0.922 & 0.978 & 0.833 \\
%      \textbf{WARP}
%        & 0.938 & 0.994 & 0.934 & 0.325 & 0.765 & 0.373 & 0.933 & 0.989 & 0.846 \\
%      \hline
%      \textbf{LSEP}
%        & 0.943 & 0.995 & 0.937 & 0.351 & 0.778 & 0.380 & 0.939 & 0.991 & 0.853 \\
%      \hline
%      \textbf{wLSEP}
%        & 0.947 & 0.995 & 0.942 & 0.362 & 0.775 & 0.380 & 0.951 & 0.992 & 0.868 \\
%      \hline
%      \bottomrule
%      \end{tabular}
%    \caption{Dataset transfer performance using ResNet-50 I3D models pretrained on Kinetics, Moments in Time (MiT) and the proposed Multi-Moments in Time dataset (M-MiT).}%
%    \label{fig:cross_dataset_image}%
%\end{table*}

\subsection{Loss Function Comparison}

\subsubsection{Results on M-MiT}
Table \ref{table:results} displays results of the models trained using different loss functions on the proposed Multi-Moments in Time dataset.  We can see that the proposed wLSEP loss function significantly outperforms the other approaches at optimization.  The combination of stable pair-wise ranking with class balancing is very effective in training our multi-label network.  Interestingly, for the TSM model BCE produced a slightly better macro mAP even though wLSEP performed better in terms of micro mAP, top-1 and top-5 accuracy.  BCE is well suited for weighted learning, as shown in Equation \ref{eq:bce}, which is likely aiding in the balanced class prediction performance here.  We want to point out that wLSEP does beat it in all of the other metrics and achieves a superior macro mAP score with the I3D model ($39.4$ compared to $37.8$ for BCE in both I3D and TSM).  This is consistent with the results from other datasets as shown in the following section.%, and  possibly due to ranking losses being better suited for datasets without full coverage as discussed in Section \ref{sec:loss}.

For our audio network the performance separation was much smaller but still results in our proposed wLSEP loss function achieving the best results.  For audio we only train and evaluate our models on videos containing audio streams.  Fusing the audio and I3D networks via an SVM results in slightly higher top-1 and top-5 accuracies than using I3D alone but does not significantly improve the performance.  We choose I3D here due to it having, on average, better performance across the metrics.  This small improvement in the fusion results is likely due to the small number of videos in the validation set that contain audio streams as well as the dominance of visual-based labels in the dataset.

\subsubsection{Comparing MiT and M-MiT Models}
\label{sec:single_to_multi}

To validate the claims that adding multiple labels improves model evaluation we compared the results of a model trained on MiT using an evaluation set of videos with their original single label, from MiT, to results using the same set of videos and model but with additional annotated labels, from M-MiT, added to each video (Table \ref{table:single_to_multi_results}).  For a fair comparison we only consider labels that are shared between MiT and M-MiT resulting in a set of 8,761 videos that we use for both single and multi-label evaluation in the table.  These results show that adding additional labels to the videos improves our ability to properly evaluate the model.  When only considering a single label for each video, the evaluation points to the model performing much worse than when we add in the additional labels from M-MiT.  This is due to not having the full set of labels in the MiT evaluation set.  We additionally present results comparing the our best M-MiT on the same set of videos and labels showing that adding additional labels in training significantly improves model results.  Both models in the table are I3D Resnet50 networks.

\subsubsection{Results on Other Datasets}

To further validate our wLSEP loss function, Table \ref{table:other_dataset_results} shows a comparison between wLSEP, LSEP, WARP and BCE on four different multi-label datasets.  This includes two image datasets for object detection (Pascal VOC \cite{Everingham10} and MS-COCO \cite{10.1007/978-3-319-10602-1_48}) as well as two video datasets for action detection (AVA \cite{gu2017ava} and MultiThumos \cite{yeung2015every}).  For the image datasets we trained a ResNet-50 and we used a ResNet-50 I3D model with 16 frames for the video datasets.  We can see in the table that the proposed wLSEP loss function consistently produces the strongest results in each dataset.

\subsection{Transfer experiments}

\begin{table*}[tb]
\centering
\setlength{\tabcolsep}{2.5pt}
\begin{tabular}{ c | c c c | c c c | c c c | c c c | c c c|}
      %\toprule       
      &  \multicolumn{15}{c}{\textbf{Fine-Tuned}} \\
       \multirow{2}{*}{\textbf{Pretrained}} & \multicolumn{3}{|c|}{\textbf{UCF-101}}  & \multicolumn{3}{|c|}{\textbf{HMDB}} & \multicolumn{3}{|c|}{\textbf{AVA}} & \multicolumn{3}{|c|}{\textbf{MultiTHUMOS}} & \multicolumn{3}{|c|}{\textbf{Charades}} \\ 
        & \textbf{Top-1}  & \textbf{Top-5} & \textbf{mAP} & \textbf{Top-1}  & \textbf{Top-5} & \textbf{mAP} & \textbf{Top-1}  & \textbf{Top-5} & \textbf{mAP} & \textbf{Top-1}  & \textbf{Top-5} & \textbf{mAP} & \textbf{Top-1}  & \textbf{Top-5} & \textbf{mAP}  \\
      \hline %\cmidrule{1-4}
      \textbf{Kinetics}
        & 0.905 & 0.987 & 0.942 & 0.726 & 0.911 & 0.809  & 0.781  &   0.973 &   0.772  & 0.837  & 0.964 & 0.821 & 0.388 & 0.731 & 0.303 \\
      \hline
      \textbf{MiT}
        & \textbf{0.908} & 0.986 & \textbf{0.943} & \textbf{0.756} & \textbf{0.937}  & \textbf{0.836}  & 0.802 & 0.976 & 0.791 & 0.849 & 0.968 & 0.838 & 0.383 & 0.711  &  0.305  \\
      \hline
      \textbf{M-MiT}
        &  0.892 &  0.980 &   0.932 & 0.740 &  0.921 &  0.819 & \textbf{0.807} &  \textbf{0.977} &    \textbf{0.795} &  \textbf{0.873} &  \textbf{0.972} &    \textbf{0.869} &  \textbf{0.414} &  \textbf{0.740} &    \textbf{0.306} \\
      \bottomrule
      \end{tabular}
    \caption{Dataset transfer performance using ResNet-50 I3D models pretrained on Kinetics, Moments in Time (MiT) and the proposed Multi-Moments in Time dataset (M-MiT).}%
    \label{fig:cross_dataset_image}%
\end{table*}

To evaluate the strength of the features learned from M-MiT we conducted a set of transfer experiments comparing ResNet-50 I3D models pretrained on Kinetics \cite{kay2017kinetics}, Moments in Time (MiT) and Multi-Moments in Time (M-MiT). We use the ResNet-50 I3D model trained with the wLSEP loss function as this model gave us the best single stream performance on our dataset (Table \ref{table:results}).  We compare our results when transferring to two single label datasets UCF-101 \cite{soomro2012ucf101} and HMDB \cite{10.1007/978-3-642-33374-3_41} as well as three multi-label datasets AVA \cite{gu2017ava}, MultiThumos \cite{yeung2015every} and Charades \cite{DBLP:journals/corr/SigurdssonVWFLG16}.  

\begin{figure}
  \centering
  \includegraphics[width=0.8\linewidth]{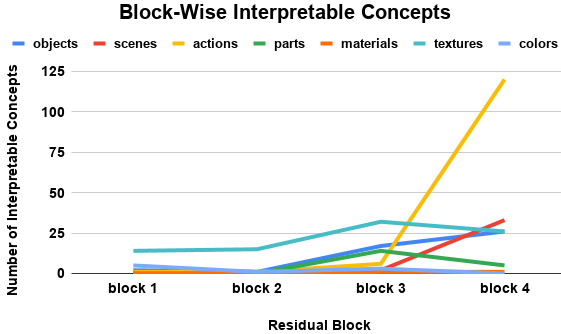}
  \caption{\textbf{ResNet block-wise interpretability} Visualize how different semantic concepts - objects, scenes and actions emerge across residual blocks of the ResNet-50 network.}
  \label{fig:blockwise}
\end{figure}

Table \ref{fig:cross_dataset_image} shows the results of the transfer task where the top-1, top-5 and mAP scores are calculated by evaluating on the validation set of the dataset used to fine-tune the model.  Here we refer to the micro mAP as mAP.  We can see from the results that pretraining on M-MiT consistently results in better performance on the multi-label datasets.  This makes sense as MiT and Kinetics are single-label datasets and were not trained to handle multiple co-occurring actions.  For the single label datasets, MiT achieves very close performance to Kinetics on UCF-101 with M-MiT following in third.  On HMDB, the M-MiT model performs a little worse than MiT and a little better than Kinetics. These results are fairly consistent with prior comparisons between Kinetics and MiT \cite{monfortmoments} and show that M-MiT pretrained models excel when transferring to multi-label settings.

\begin{figure*}
\centering
\begin{minipage}{0.7\textwidth}
\centering
\begin{subfigure}{.19\textwidth}
  \centering
  \includegraphics[width=\linewidth]{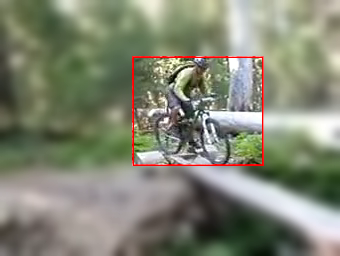}
  \caption*{bicycling}
\end{subfigure}
\begin{subfigure}{.19\textwidth}
  \centering
  \includegraphics[width=\linewidth]{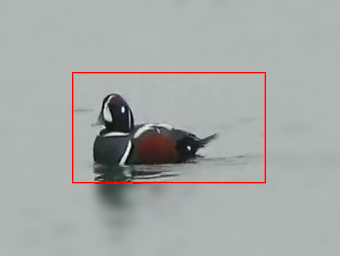}
  \caption*{floating}
\end{subfigure}
\begin{subfigure}{.19\textwidth}
  \centering
  \includegraphics[width=\linewidth]{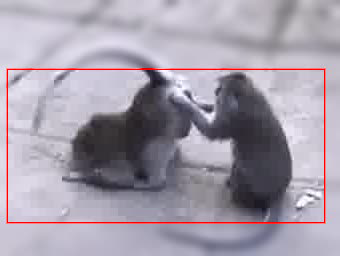}
  \caption*{grooming}
\end{subfigure}
\begin{subfigure}{.19\textwidth}
  \centering
  \includegraphics[width=\linewidth]{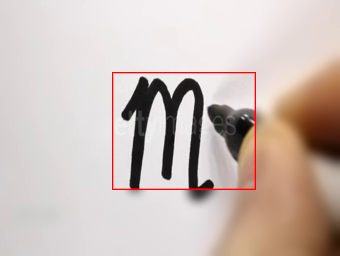}
  \caption*{writing}
\end{subfigure}
\begin{subfigure}{.19\textwidth}
  \centering
  \includegraphics[width=\linewidth]{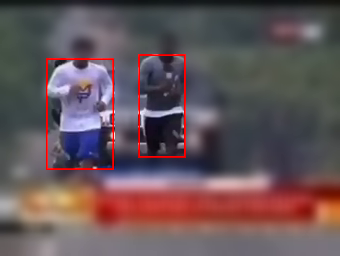}
  \caption*{jogging}
\end{subfigure}
\vspace{5pt}
\caption{\textbf{Localized action regions:} Bounding boxes annotated around 549 different action categories in 67,468 image frames each selected from unique videos in the Moments in Time, Kinetics, and Something-Something datasets.}
\label{fig:action_regions}
\end{minipage}

\vspace{10pt}

\begin{minipage}{\textwidth}
\centering
\includegraphics[width=\linewidth]{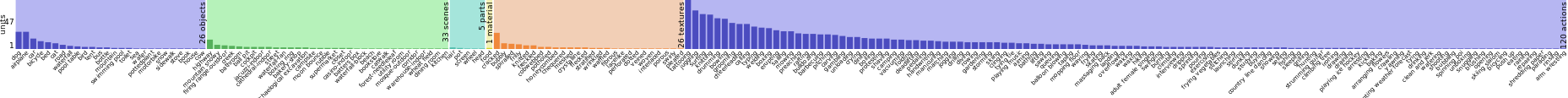}
\caption{\textbf{Graph of learned action concepts ordered by the number of features associated with each concept.}}
\label{fig:concept_graph}
\end{minipage}

\vspace{10pt}

\begin{minipage}{\textwidth}
\centering
\begin{minipage}{0.33\textwidth}
\caption*{\textbf{burning}}
\begin{subfigure}{0.49\textwidth}
\includegraphics[width=\linewidth]{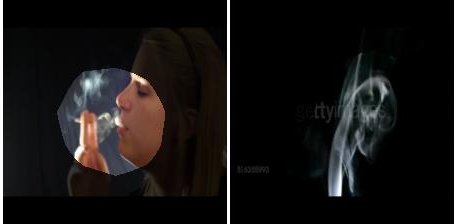}
\caption*{Unit 684}
\end{subfigure}
\begin{subfigure}{0.49\textwidth}
\includegraphics[width=\linewidth]{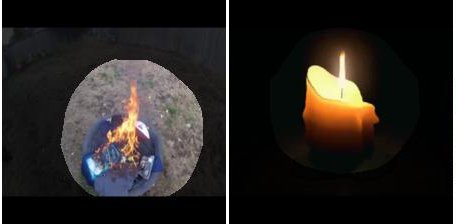}
\caption*{Unit 1417}
\end{subfigure}
\end{minipage}
\begin{minipage}{0.33\textwidth}
\caption*{\textbf{climbing}}
\begin{subfigure}{0.49\textwidth}
\includegraphics[width=\linewidth]{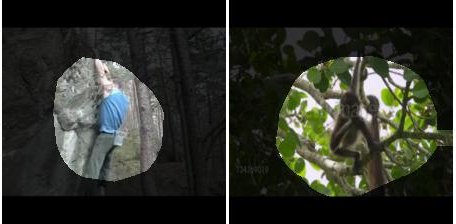}
\caption*{Unit 925}
\end{subfigure}
\begin{subfigure}{0.49\textwidth}
\includegraphics[width=\linewidth]{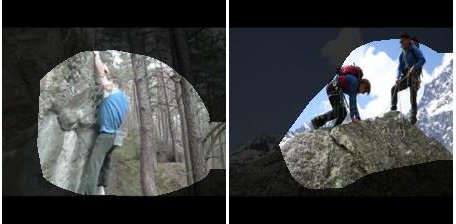}
\caption*{Unit 1858}
\end{subfigure}
\end{minipage}
\begin{minipage}{0.33\textwidth}
\caption*{\textbf{crawling}}
\begin{subfigure}{0.49\textwidth}
\includegraphics[width=\linewidth]{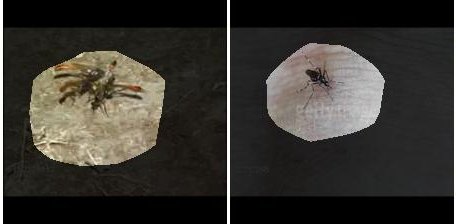}
\caption*{Unit 617}
\end{subfigure}
\begin{subfigure}{0.49\textwidth}
\includegraphics[width=\linewidth]{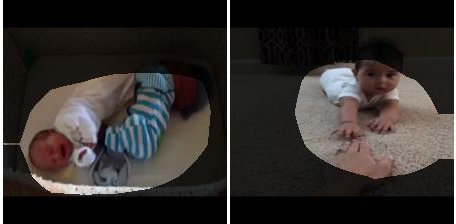}
\caption*{Unit 828}
\end{subfigure}
\end{minipage}
\begin{minipage}{0.33\textwidth}
\caption*{\textbf{drinking}}
\begin{subfigure}{0.49\textwidth}
\includegraphics[width=\linewidth]{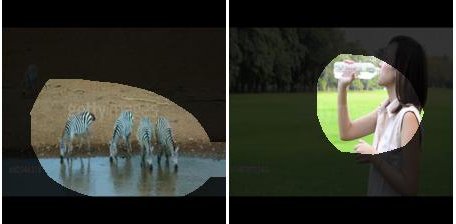}
\caption*{Unit 823}
\end{subfigure}
\begin{subfigure}{0.49\textwidth}
\includegraphics[width=\linewidth]{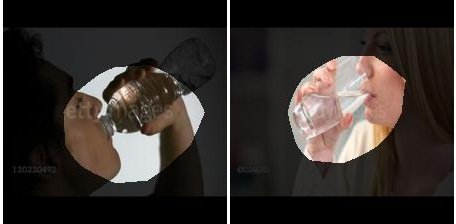}
\caption*{Unit 1321}
\end{subfigure}
\end{minipage}
\begin{minipage}{0.33\textwidth}
\caption*{\textbf{folding}}
\begin{subfigure}{0.49\textwidth}
\includegraphics[width=\linewidth]{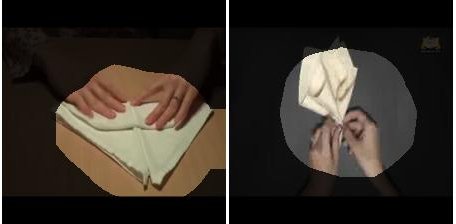}
\caption*{Unit 1699}
\end{subfigure}
\begin{subfigure}{0.49\textwidth}
\includegraphics[width=\linewidth]{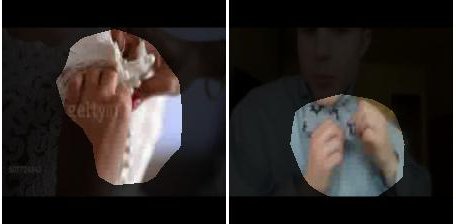}
\caption*{Unit 2032}
\end{subfigure}
\end{minipage}
\begin{minipage}{0.33\textwidth}
\caption*{\textbf{pouring}}
\begin{subfigure}{0.49\textwidth}
\includegraphics[width=\linewidth]{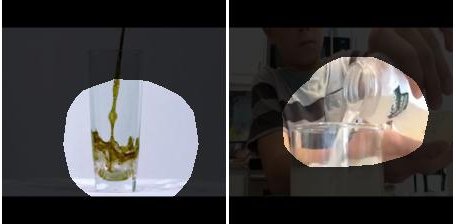}
\caption*{Unit 21}
\end{subfigure}
\begin{subfigure}{0.49\textwidth}
\includegraphics[width=\linewidth]{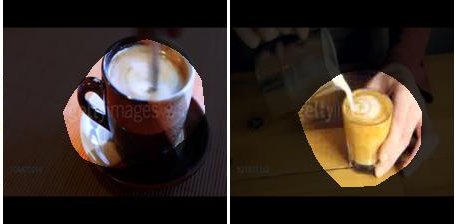}
\caption*{Unit 821}
\end{subfigure}
\end{minipage}
\caption{\textbf{Visualization of learned action concepts:}  Different features learn different representations of the same action.  For example, units 684 and 1417 can both be interpreted as learning the concept of \emph{burning} (top left).  However, unit 684 learns to correlate \emph{smoke} with the action while unit 1417 correlates it with a \emph{flame}.}
\label{fig:learned_concepts}
\end{minipage}

\vspace{10pt}

\begin{minipage}{\linewidth}
\centering
\begin{minipage}{0.33\textwidth}
\caption*{\textbf{Unit 626:}  \emph{water} (material) $\rightarrow$ \emph{rowing}}
\begin{subfigure}{\textwidth}
\includegraphics[width=\linewidth]{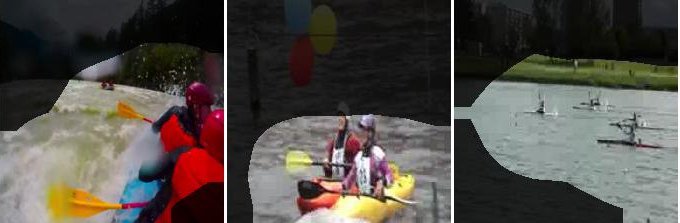}
\end{subfigure}
\caption*{}
\end{minipage}
\begin{minipage}{0.33\textwidth}
\caption*{\textbf{Unit 2025:}  \emph{bubbly} (texture) $\rightarrow$ \emph{surfing}}
\begin{subfigure}{\textwidth}
\includegraphics[width=\linewidth]{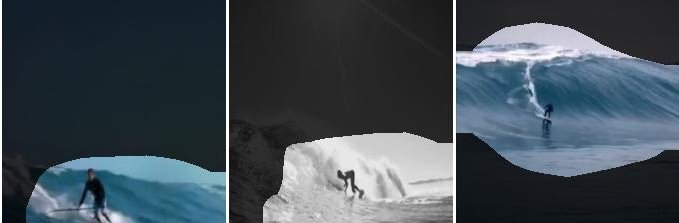}
\end{subfigure}
\caption*{}
\end{minipage}
\begin{minipage}{0.33\textwidth}
\caption*{\textbf{Unit 1944:}  \emph{corridor} (scene) $\rightarrow$ \emph{bowling}}
\begin{subfigure}{\textwidth}
\includegraphics[width=\linewidth]{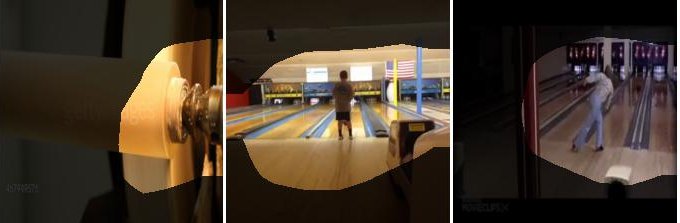}
\end{subfigure}
\caption*{}
\end{minipage}
\caption{\textbf{Improved feature interpretation:} Examples of the 953 units from the final residual block of a ResNet-50 that changed their main interpretation when actions were added to the Broden dataset.}
\label{fig:broden_to_action}
\end{minipage}
\end{figure*}

\section{Model Analysis}

\subsection{Multi-Label Class Activation Mapping}
\label{sec:multiCamResults}

In Figure \ref{fig:multi_cam} we see an example of applying our multi-label CAM filter to an image with two actions.  Using standard CAM filters it is difficult to disambiguate unique regions used for the different actions detected in the image (\emph{coaching} and \emph{punching}).  As with many actions, the image regions used by the model for detection are strongly correlated.  We use the proposed multi-CAM approach described in Section \ref{sec:cam} to highlight the differences between the CAM results for each predicted class.  These differences help us visualize how the model has learned to discriminate different classes that may be present in the same image.  This is especially useful for classes that may share multiple visual characteristics.  In Figure \ref{fig:multi_cam_collage} we include a diverse set of examples showing results of this method for different action combinations.

\subsection{Learned Action Interpretation}
\label{sec:interpretationResults}

Using the approach described in Section \ref{sec:netdissect} we are able to identify a total of 211 concepts consisting of 1 material, 5 part, 26 texture, 26 object, 33 scene and 120 action concepts learned in 2023 different features out of 2048 (Figure \ref{fig:concept_graph}) units in the final convolutional layer (block4) of a ResNet-50 network trained on the Multi-Moments in Time dataset. Figure \ref{fig:learned_concepts} highlights some of the learned concepts. For example, the network associates \emph{crawling} with babies as many of our videos of crawling typically depict babies \emph{crawling}. These are the types of data and class biases that are useful to identify via network interpretation that may have gone unnoticed without the ability to identify action concepts.

Table \ref{fig:broden_action_comparison} highlights the fact that including actions in the Broden dataset helps to interpret a much larger portion of the features in block 4 of a ResNet-50 trained for action recognition. With the original Broden set (no actions) NetDissect identified 185 concepts in 1351/2048 features. Adding actions to this set allowed us to identify 2023/2048 features that can be interpreted for 211 different concepts including 120 actions. This jump in the number of interpretable features makes sense for the final block of a model trained for action recognition and suggests that excluding action concepts misses a large portion of useful information when interpreting these models.

The results from the combined set highlight that some of the features previously interpreted by the original Broden set as object or texture concepts are more closely aligned with actions. A few examples of this behavior, and how adding actions to the Broden dataset improves our interpretation, can be seen in Figure \ref{fig:broden_to_action}.  For example, unit, or feature, 2025 was previously interpreted to be most closely associated with the texture "bubbly", but after adding the new action regions to the Broden set we found that the feature is actually more strongly correlated with the action "surfing".

%For example, unit 13 was identified %using the Broden set 
%as learning the concept \emph{potted plant} with an IoU of 0.06, but if we include action concepts the unit is found to be more correlated with the action \emph{gardening} with an IoU of 0.15.  Similarly, unit 290 was identified by Broden as learning the texture concept \emph{cracked} with an IoU of 0.1 and including actions we found a greater association with the action \emph{typing} with an IoU of 0.34.  Features for identifying the ridges between the keys in the keyboards commonly found in actions of \emph{typing} were correctly activating for the texture \emph{cracked}.  

Examining these interpretable features and how they contribute to a networks output allows us to build a better understanding of how our models will function when presented with different data.  This understanding can be an important tool for model design.  For example, improved understanding of model interpretation has recently been shown to be useful in improving adversarial robustness \cite{pmlr-v119-boopathy20a}, obtaining state-of-the-art autoencoder-based generative models \cite{Esser_2020_CVPR} and manipulating the output of a GAN \cite{bau2019gandissect}.

\begin{table}[tb]
\centering
\begin{tabular}{ c | c | c}
       \multirow{2}*{\textbf{Category}} & \multirow{2}*{Concepts}  & Interpretable \\ 
        &   & Features \\ 
      \cmidrule{1-3}
      Broden & 185 & 1351 \\
      Action Regions & 140 & 1995 \\
      Broden+Action Regions & 211 & 2023 \\
      \bottomrule
      \end{tabular}
    \caption{Comparison of the number of concepts and interpretable features identified by NetDissect given the Broden dataset, the Action Region dataset and the combined dataset on block 4 of a ResNet-50 trained for action recognition.}%
    \label{fig:broden_action_comparison}
\end{table}

\subsubsection{Block-wise Interpretability}

To understand how individual units evolve over residual blocks we evaluate the interpretability of features from different blocks of a ResNet-50 network trained for action recognition on the Moments in Time dataset \cite{monfortmoments} in terms of objects, scenes, actions and textures. In Figure \ref{fig:blockwise} we observe that action features mainly emerge in the last convolutional block (block 4) of the model. It is interesting to note that object and scene features are learned even if the model is not explicitly trained to recognize objects or scenes suggesting that object and scene recognition aids action classification.

\subsubsection{Interpretable feature relationships}

Examining these interpretable features and how they contribute to a networks output allows us to build a better understanding of how our models will function when presented with different data.  For example, we can consider a feature in the final residual block of the network that has been interpreted as a \emph{highway} scene feature.  If we activate only this unit by setting its values to 1 and all other feature (including bias) values to 0 %and examine the output 
we can identify which actions are correlated with the fact that a video may take place on a \emph{highway}.  In this case the actions that achieve the highest output are \emph{hitchhiking}, \emph{towing}, \emph{swerving}, \emph{riding}, and \emph{driving}.  These interpretable feature-class relationships make sense as all of these actions are likely to occur near a \emph{highway}.

\section{Conclusion}

Progress in the field of video understanding will come from many fronts, including training our models with richer and more complete information, so they can start achieving recognition performances in par with humans. Augmenting a large-scale dataset by doubling the number of activity labels,  we present baseline results on the Multi-Moments dataset as well as improved methods for visualizing and interpreting models trained for multi-label action detection.

\textbf{Acknowledgements:}
This work was supported by the MIT-IBM Watson AI Lab and its member companies, Nexplore and Woodside, Google faculty award  and SystemsThatLearn@CSAIL award (to A.O), as well as the Intelligence Advanced Research Projects Activity (IARPA) via Department of Interior/ Interior Business Center (DOI/IBC) contract number D17PC00341. The U.S. Government is authorized to reproduce and distribute reprints for Governmental purposes notwithstanding any copyright annotation thereon.  Disclaimer: The views and conclusions contained herein are those of the authors and should not be interpreted as necessarily representing the official policies or endorsements, either expressed or implied, of IARPA, DOI/IBC, or the U.S. Government.  Special thanks to David Bau for assisting in extending the Broden Dataset for Network Dissection and Allen Lee for aiding in the design of our demo.

{
\balance
\footnotesize
\bibliographystyle{ieee}
\bibliography{main}

\begin{thebibliography}{10}\itemsep=-1pt

\bibitem{youtube8m}
S.~Abu{-}El{-}Haija, N.~Kothari, J.~Lee, P.~Natsev, G.~Toderici,
  B.~Varadarajan, and S.~Vijayanarasimhan.
\newblock Youtube-8m: {A} large-scale video classification benchmark.
\newblock {\em CoRR}, abs/1609.08675, 2016.

\bibitem{NIPS2016_6146}
Y.~Aytar, C.~Vondrick, and A.~Torralba.
\newblock Soundnet: Learning sound representations from unlabeled video.
\newblock In D.~D. Lee, M.~Sugiyama, U.~V. Luxburg, I.~Guyon, and R.~Garnett,
  editors, {\em Advances in Neural Information Processing Systems 29}. 2016.

\bibitem{netdissect2017}
D.~Bau, B.~Zhou, A.~Khosla, A.~Oliva, and A.~Torralba.
\newblock Network dissection: Quantifying interpretability of deep visual
  representations.
\newblock In {\em Computer Vision and Pattern Recognition}, 2017.

\bibitem{Bau_2017_CVPR}
D.~Bau, B.~Zhou, A.~Khosla, A.~Oliva, and A.~Torralba.
\newblock Network dissection: Quantifying interpretability of deep visual
  representations.
\newblock In {\em Proceedings of the IEEE Conference on Computer Vision and
  Pattern Recognition (CVPR)}, July 2017.

\bibitem{bau2019gandissect}
D.~Bau, J.-Y. Zhu, H.~Strobelt, B.~Zhou, J.~B. Tenenbaum, W.~T. Freeman, and
  A.~Torralba.
\newblock Gan dissection: Visualizing and understanding generative adversarial
  networks.
\newblock In {\em Proceedings of the International Conference on Learning
  Representations (ICLR)}, 2019.

\bibitem{bell13opensurfaces}
S.~Bell, P.~Upchurch, N.~Snavely, and K.~Bala.
\newblock Open{S}urfaces: A richly annotated catalog of surface appearance.
\newblock {\em ACM Trans. on Graphics (SIGGRAPH)}, 32(4), 2013.

\bibitem{pmlr-v119-boopathy20a}
A.~Boopathy, S.~Liu, G.~Zhang, C.~Liu, P.-Y. Chen, S.~Chang, and L.~Daniel.
\newblock Proper network interpretability helps adversarial robustness in
  classification.
\newblock In H.~D. III and A.~Singh, editors, {\em Proceedings of the 37th
  International Conference on Machine Learning}, volume 119 of {\em Proceedings
  of Machine Learning Research}, pages 1014--1023. PMLR, 13--18 Jul 2020.

\bibitem{caba2015activitynet}
F.~Caba~Heilbron, V.~Escorcia, B.~Ghanem, and J.~Carlos~Niebles.
\newblock Activitynet: A large-scale video benchmark for human activity
  understanding.
\newblock In {\em CVPR}, 2015.

\bibitem{carreira2017quo}
J.~Carreira and A.~Zisserman.
\newblock Quo vadis, action recognition? a new model and the kinetics dataset.
\newblock {\em Proc. ICCV}, 2017.

\bibitem{chen_cvpr14}
X.~Chen, R.~Mottaghi, X.~Liu, S.~Fidler, R.~Urtasun, and A.~Yuille.
\newblock Detect what you can: Detecting and representing objects using
  holistic models and body parts.
\newblock In {\em IEEE Conference on Computer Vision and Pattern Recognition
  (CVPR)}, 2014.

\bibitem{Damen2018EPICKITCHENS}
D.~Damen, H.~Doughty, G.~M. Farinella, S.~Fidler, A.~Furnari, E.~Kazakos,
  D.~Moltisanti, J.~Munro, T.~Perrett, W.~Price, and M.~Wray.
\newblock Scaling egocentric vision: The epic-kitchens dataset.
\newblock In {\em European Conference on Computer Vision (ECCV)}, 2018.

\bibitem{deng2009imagenet}
J.~Deng, W.~Dong, R.~Socher, L.-J. Li, K.~Li, and L.~Fei-Fei.
\newblock Imagenet: A large-scale hierarchical image database.
\newblock In {\em CVPR}, 2009.

\bibitem{Elisseeff:2001:KMM:2980539.2980628}
A.~Elisseeff and J.~Weston.
\newblock A kernel method for multi-labelled classification.
\newblock In {\em Proceedings of the 14th International Conference on Neural
  Information Processing Systems: Natural and Synthetic}, NIPS'01, pages
  681--687, Cambridge, MA, USA, 2001. MIT Press.

\bibitem{Esser_2020_CVPR}
P.~Esser, R.~Rombach, and B.~Ommer.
\newblock A disentangling invertible interpretation network for explaining
  latent representations.
\newblock In {\em Proceedings of the IEEE/CVF Conference on Computer Vision and
  Pattern Recognition (CVPR)}, June 2020.

\bibitem{Everingham10}
M.~Everingham, L.~Van~Gool, C.~K.~I. Williams, J.~Winn, and A.~Zisserman.
\newblock The pascal visual object classes (voc) challenge.
\newblock {\em International Journal of Computer Vision}, 88(2):303--338, June
  2010.

\bibitem{gong2013deep}
Y.~Gong, Y.~Jia, T.~Leung, A.~Toshev, and S.~Ioffe.
\newblock Deep convolutional ranking for multilabel image annotation.
\newblock {\em arXiv preprint arXiv:1312.4894}, 2013.

\bibitem{goyal2017something}
R.~Goyal, S.~Kahou, V.~Michalski, J.~Materzy{\'n}ska, S.~Westphal, H.~Kim,
  V.~Haenel, I.~Fruend, P.~Yianilos, M.~Mueller-Freitag, et~al.
\newblock The" something something" video database for learning and evaluating
  visual common sense.
\newblock {\em arXiv preprint arXiv:1706.04261}, 2017.

\bibitem{gu2017ava}
C.~Gu, C.~Sun, S.~Vijayanarasimhan, C.~Pantofaru, D.~A. Ross, G.~Toderici,
  Y.~Li, S.~Ricco, R.~Sukthankar, C.~Schmid, et~al.
\newblock Ava: A video dataset of spatio-temporally localized atomic visual
  actions.
\newblock {\em arXiv preprint arXiv:1705.08421}, 2017.

\bibitem{jiang2014thumos}
Y.~Jiang, J.~Liu, A.~R. Zamir, G.~Toderici, I.~Laptev, M.~Shah, and
  R.~Sukthankar.
\newblock Thumos challenge: Action recognition with a large number of classes,
  2014.

\bibitem{kay2017kinetics}
W.~Kay, J.~Carreira, K.~Simonyan, B.~Zhang, C.~Hillier, S.~Vijayanarasimhan,
  F.~Viola, T.~Green, T.~Back, P.~Natsev, et~al.
\newblock The kinetics human action video dataset.
\newblock {\em arXiv preprint arXiv:1705.06950}, 2017.

\bibitem{10.1007/978-3-642-33374-3_41}
H.~Kuehne, H.~Jhuang, R.~Stiefelhagen, and T.~Serre.
\newblock Hmdb51: A large video database for human motion recognition.
\newblock In W.~E. Nagel, D.~H. Kr{\"o}ner, and M.~M. Resch, editors, {\em High
  Performance Computing in Science and Engineering `12}, Berlin, Heidelberg,
  2013. Springer Berlin Heidelberg.

\bibitem{Li2017ImprovingPR}
Y.~Li, Y.~Song, and J.~Luo.
\newblock Improving pairwise ranking for multi-label image classification.
\newblock {\em 2017 IEEE Conference on Computer Vision and Pattern Recognition
  (CVPR)}, pages 1837--1845, 2017.

\bibitem{Lin_2019_ICCV}
J.~Lin, C.~Gan, and S.~Han.
\newblock Tsm: Temporal shift module for efficient video understanding.
\newblock In {\em The IEEE International Conference on Computer Vision (ICCV)},
  October 2019.

\bibitem{10.1007/978-3-319-10602-1_48}
T.-Y. Lin, M.~Maire, S.~Belongie, J.~Hays, P.~Perona, D.~Ramanan,
  P.~Doll{\'a}r, and C.~L. Zitnick.
\newblock Microsoft coco: Common objects in context.
\newblock In D.~Fleet, T.~Pajdla, B.~Schiele, and T.~Tuytelaars, editors, {\em
  Computer Vision -- ECCV 2014}, pages 740--755, Cham, 2014. Springer
  International Publishing.

\bibitem{DBLP:journals/corr/abs-1301-3781}
T.~Mikolov, K.~Chen, G.~Corrado, and J.~Dean.
\newblock Efficient estimation of word representations in vector space.
\newblock {\em CoRR}, abs/1301.3781, 2013.

\bibitem{Miller:1995:WLD:219717.219748}
G.~A. Miller.
\newblock Wordnet: A lexical database for english.
\newblock {\em Commun. ACM}, 38(11):39--41, Nov. 1995.

\bibitem{monfortmoments}
M.~Monfort, A.~Andonian, B.~Zhou, K.~Ramakrishnan, S.~A. Bargal, T.~Yan,
  L.~Brown, Q.~Fan, D.~Gutfruend, C.~Vondrick, et~al.
\newblock Moments in time dataset: one million videos for event understanding.
\newblock {\em IEEE Transactions on Pattern Analysis and Machine Intelligence},
  pages 1--8, 2019.

\bibitem{mottaghi_cvpr14}
R.~Mottaghi, X.~Chen, X.~Liu, N.-G. Cho, S.-W. Lee, S.~Fidler, R.~Urtasun, and
  A.~Yuille.
\newblock The role of context for object detection and semantic segmentation in
  the wild.
\newblock In {\em IEEE Conference on Computer Vision and Pattern Recognition
  (CVPR)}, 2014.

\bibitem{DBLP:journals/corr/SigurdssonVWFLG16}
G.~A. Sigurdsson, G.~Varol, X.~Wang, A.~Farhadi, I.~Laptev, and A.~Gupta.
\newblock Hollywood in homes: Crowdsourcing data collection for activity
  understanding.
\newblock {\em CoRR}, abs/1604.01753, 2016.

\bibitem{simonyan2014two}
K.~Simonyan and A.~Zisserman.
\newblock Two-stream convolutional networks for action recognition in videos.
\newblock In {\em Advances in neural information processing systems}, pages
  568--576, 2014.

\bibitem{soomro2012ucf101}
K.~Soomro, A.~R. Zamir, and M.~Shah.
\newblock Ucf101: A dataset of 101 human actions classes from videos in the
  wild.
\newblock {\em arXiv preprint arXiv:1212.0402}, 2012.

\bibitem{tran2015learning}
D.~Tran, L.~Bourdev, R.~Fergus, L.~Torresani, and M.~Paluri.
\newblock Learning spatiotemporal features with 3d convolutional networks.
\newblock In {\em CVPR}, 2015.

\bibitem{wang2016cnn}
J.~Wang, Y.~Yang, J.~Mao, Z.~Huang, C.~Huang, and W.~Xu.
\newblock Cnn-rnn: A unified framework for multi-label image classification.
\newblock In {\em Computer Vision and Pattern Recognition (CVPR), 2016 IEEE
  Conference on}, pages 2285--2294. IEEE, 2016.

\bibitem{Weston:2011:WSU:2283696.2283856}
J.~Weston, S.~Bengio, and N.~Usunier.
\newblock Wsabie: Scaling up to large vocabulary image annotation.
\newblock In {\em Proceedings of the Twenty-Second International Joint
  Conference on Artificial Intelligence - Volume Volume Three}, IJCAI'11, pages
  2764--2770. AAAI Press, 2011.

\bibitem{yang2016exploit}
H.~Yang, J.~Tianyi~Zhou, Y.~Zhang, B.-B. Gao, J.~Wu, and J.~Cai.
\newblock Exploit bounding box annotations for multi-label object recognition.
\newblock In {\em Proceedings of the IEEE Conference on Computer Vision and
  Pattern Recognition}, pages 280--288, 2016.

\bibitem{yeung2015every}
S.~Yeung, O.~Russakovsky, N.~Jin, M.~Andriluka, G.~Mori, and L.~Fei-Fei.
\newblock Every moment counts: Dense detailed labeling of actions in complex
  videos.
\newblock {\em arXiv preprint arXiv:1507.05738}, 2015.

\bibitem{zhang2018multi}
J.~Zhang, Q.~Wu, C.~Shen, J.~Zhang, and J.~Lu.
\newblock Multi-label image classification with regional latent semantic
  dependencies.
\newblock {\em IEEE Transactions on Multimedia}, 2018.

\bibitem{Zhang:2006:MNN:1159162.1159294}
M.-L. Zhang and Z.-H. Zhou.
\newblock Multilabel neural networks with applications to functional genomics
  and text categorization.
\newblock {\em IEEE Trans. on Knowl. and Data Eng.}, 18(10):1338--1351, Oct.
  2006.

\bibitem{8417924}
B.~Zhou, D.~Bau, A.~Oliva, and A.~Torralba.
\newblock Interpreting deep visual representations via network dissection.
\newblock {\em IEEE Transactions on Pattern Analysis and Machine Intelligence},
  pages 1--1, 2018.

\bibitem{zhou2015cnnlocalization}
B.~Zhou, A.~Khosla, L.~A., A.~Oliva, and A.~Torralba.
\newblock {Learning Deep Features for Discriminative Localization.}
\newblock {\em CVPR}, 2016.

\bibitem{zhou2016learning}
B.~Zhou, A.~Khosla, A.~Lapedriza, A.~Oliva, and A.~Torralba.
\newblock Learning deep features for discriminative localization.
\newblock In {\em Proceedings of the IEEE Conference on Computer Vision and
  Pattern Recognition}, pages 2921--2929, 2016.

\bibitem{zhou2016semantic}
B.~Zhou, H.~Zhao, X.~Puig, S.~Fidler, A.~Barriuso, and A.~Torralba.
\newblock Semantic understanding of scenes through the ade20k dataset.
\newblock {\em arXiv preprint arXiv:1608.05442}, 2016.

\end{thebibliography}
}

\end{document}